\documentclass[10pt,twocolumn,letterpaper]{article}

\usepackage{iccv}
\usepackage{times}
\usepackage{epsfig}
\usepackage{graphicx}
\usepackage{amsmath}
\usepackage{amssymb}
\usepackage{amssymb}
\usepackage{url}
\usepackage{booktabs}
\usepackage{multirow}
\usepackage{subfigure}
\usepackage{lib}
\usepackage{makecell}
\usepackage{booktabs}
\usepackage[pagebackref=true,breaklinks=true,letterpaper=true,colorlinks,bookmarks=false]{hyperref}

\iccvfinalcopy % *** Uncomment this line for the final submission

\def\iccvPaperID{1698} % *** Enter the ICCV Paper ID here

\newcommand{\X}{$\times$\xspace}

\newcommand{\vggsixteen}{VGG16\xspace}
\newcommand{\roi}{RoI\xspace}
\newcommand{\Sm}{{\bf S}\xspace}
\newcommand{\Med}{{\bf M}\xspace}
\newcommand{\Lg}{{\bf L}\xspace}
\newcommand{\ZF}{{\bf ZF}\xspace}

\newcolumntype{x}{>\small c}
\newcolumntype{L}[1]{>{\raggedright\let\newline\\\arraybackslash\hspace{0pt}}m{#1}}
\newcolumntype{C}[1]{>{\centering\let\newline\\\arraybackslash\hspace{0pt}}m{#1}}
\newcolumntype{R}[1]{>{\raggedleft\let\newline\\\arraybackslash\hspace{0pt}}m{#1}}
\begin{document}

%%%%%%%%% TITLE
\title{Fast R-CNN}

\author{Ross Girshick\\
Microsoft Research\\
{\tt\small rbg@microsoft.com}
}

\maketitle
\thispagestyle{empty}

\begin{abstract}
This paper proposes a Fast Region-based Convolutional Network method \emph{(Fast R-CNN)} for object detection.
Fast R-CNN builds on previous work to efficiently classify object proposals using deep convolutional networks.
Compared to previous work, Fast R-CNN employs several innovations to improve training and testing speed while also increasing detection accuracy.
Fast R-CNN trains the very deep \vggsixteen network 9\X faster than R-CNN, is 213\X faster at test-time, and achieves a higher mAP on PASCAL VOC 2012.
Compared to SPPnet, Fast R-CNN trains \vggsixteen 3\X faster, tests 10\X faster, and is more accurate.
Fast R-CNN is implemented in Python and C++ (using Caffe) and is available under the open-source MIT License at \url{https://github.com/rbgirshick/fast-rcnn}.
\end{abstract}

\section{Introduction}

Recently, deep ConvNets \cite{krizhevsky2012imagenet,lecun89e} have significantly improved image classification \cite{krizhevsky2012imagenet} and object detection \cite{girshick2014rcnn,overfeat} accuracy.
Compared to image classification, object detection is a more challenging task that requires more complex methods to solve.
Due to this complexity, current approaches (\eg, \cite{girshick2014rcnn,he2014spp,overfeat,Zhu2015segDeepM}) train models in multi-stage pipelines that are slow and inelegant.

Complexity arises because detection requires the accurate localization of objects, creating two primary challenges.
First, numerous candidate object locations (often called ``proposals'') must be processed.
Second, these candidates provide only rough localization that must be refined to achieve precise localization.
Solutions to these problems often compromise speed, accuracy, or simplicity.
%In order to make ConvNet-based detection efficient, convolutional features are shared between candidates \cite{he2014spp,overfeat,lecun94}.
%Candidate location refinement is often implemented as a post-hoc regression of true object bounds given noisy candidates \cite{girshick2014rcnn,overfeat}.
%While \cite{he2014spp} shares features, the convolutional layers of the network remain fixed during training, which limits accuracy.
%For refinement, previous processes \cite{girshick2014rcnn,overfeat} use a separate learning stage, increasing training complexity.

In this paper, we streamline the training process for state-of-the-art ConvNet-based object detectors \cite{girshick2014rcnn,he2014spp}.
We propose a single-stage training algorithm that jointly learns to classify object proposals and refine their spatial locations.
%Our method for sharing convolutional features during training allows for the full back-propagation of errors through the network, leading to increased accuracy relative to previous approaches \cite{he2014spp}.
%Several improvements are also made to increase runtime efficiency, such as the use of truncated SVD, which is particularly effective for object detection.

The resulting method can train a very deep detection network (\vggsixteen \cite{simonyan2015verydeep}) 9\X faster than R-CNN \cite{girshick2014rcnn} and 3\X faster than SPPnet \cite{he2014spp}.
At runtime, the detection network processes images in 0.3s (excluding object proposal time) while achieving top accuracy on PASCAL VOC 2012 \cite{Pascal-IJCV} with a mAP of 66\% (vs. 62\% for R-CNN).\footnote{All timings use one Nvidia K40 GPU overclocked to 875 MHz.}

%In this paper, we make R-CNN \cite{girshick2014rcnn} \emph{clean and fast}.
%We unify training with a multi-task loss, removing unnecessary steps (separate SVM and bounding-box regression training).
%Our training algorithm updates \emph{all} network layers unlike previous R-CNN acceleration methods (SPPnet \cite{he2014spp}).
%Our system trains \vggsixteen \cite{simonyan2015verydeep} for PASCAL VOC \cite{PASCAL-ijcv} detection in 9.5 hours (9x faster than R-CNN), processes images in 0.22s ($>$200x faster than R-CNN), and achieves higher mAP than R-CNN (66\% vs. 62\% mAP on VOC12 test).

\subsection{R-CNN and SPPnet}
The Region-based Convolutional Network method (R-CNN) \cite{girshick2014rcnn} achieves excellent object detection accuracy by using a deep ConvNet to classify object proposals.
R-CNN, however, has notable drawbacks:
\begin{enumerate}
\itemsep0em
\item {\bf Training is a multi-stage pipeline.}
R-CNN first fine-tunes a ConvNet on object proposals using log loss.
Then, it fits SVMs to ConvNet features.
These SVMs act as object detectors, replacing the softmax classifier learnt by fine-tuning.
In the third training stage, bounding-box regressors are learned.
\item {\bf Training is expensive in space and time.}
For SVM and bounding-box regressor training, features are extracted from each object proposal in each image and written to disk.
With very deep networks, such as \vggsixteen, this process takes 2.5 GPU-days for the 5k images of the VOC07 trainval set.
These features require hundreds of gigabytes of storage.
%These features are cached to disk for later use in SVM and regressor training.
%The cached features occupy hundreds of gigabytes and training can suffer from high IO latency when using network disks.
\item {\bf Object detection is slow.}
At test-time, features are extracted from each object proposal in each test image.
Detection with \vggsixteen takes 47s / image (on a GPU).
%(K40 GPU, excludes object proposal time).
\end{enumerate}

R-CNN is slow because it performs a ConvNet forward pass for each object proposal, without sharing computation.
Spatial pyramid pooling networks (SPPnets) \cite{he2014spp} were proposed to speed up R-CNN by sharing computation.
%In the SPPnet method, computation is reorganized so that all object proposals share a significant portion of the overall computation.
The SPPnet method computes a convolutional feature map for the entire input image and then classifies each object proposal using a feature vector extracted from the shared feature map.
Features are extracted for a proposal by max-pooling the portion of the feature map inside the proposal into a fixed-size output (\eg, $6 \times 6$).
Multiple output sizes are pooled and then concatenated as in spatial pyramid pooling \cite{Lazebnik2006}.
SPPnet accelerates R-CNN by 10 to 100\X at test time.
Training time is also reduced by 3\X due to faster proposal feature extraction.

SPPnet also has notable drawbacks.
Like R-CNN, training is a multi-stage pipeline that involves extracting features, fine-tuning a network with log loss, training SVMs, and finally fitting bounding-box regressors.
Features are also written to disk.
%since they are used twice, once for training SVMs and again for training bounding-box regression.
But unlike R-CNN, the fine-tuning algorithm proposed in \cite{he2014spp} cannot update the convolutional layers that precede the spatial pyramid pooling.
Unsurprisingly, this limitation (fixed convolutional layers) limits the accuracy of very deep networks.

\subsection{Contributions}
We propose a new training algorithm that fixes the disadvantages of R-CNN and SPPnet, while improving on their speed and accuracy.
We call this method \emph{Fast R-CNN} because it's comparatively fast to train and test.
The Fast R-CNN method has several advantages:
\begin{enumerate}
  \itemsep0em
  \item Higher detection quality (mAP) than R-CNN, SPPnet
  \item Training is single-stage, using a multi-task loss
  \item Training can update all network layers
  \item No disk storage is required for feature caching
\end{enumerate}

Fast R-CNN is written in Python and C++ (Caffe \cite{jia2014caffe}) and is available under the open-source MIT License at \url{https://github.com/rbgirshick/fast-rcnn}.

\section{Fast R-CNN architecture and training}

\figref{arch} illustrates the Fast R-CNN architecture.
A Fast R-CNN network takes as input an entire image and a set of object proposals.
The network first processes the whole image with several convolutional (\emph{conv}) and max pooling layers to produce a conv feature map.
Then, for each object proposal a region of interest (\emph{\roi}) pooling layer extracts a fixed-length feature vector from the feature map.
Each feature vector is fed into a sequence of fully connected (\emph{fc}) layers that finally branch into two sibling output layers: one that produces softmax probability estimates over $K$ object classes plus a catch-all ``background'' class and another layer that outputs four real-valued numbers for each of the $K$ object classes.
Each set of $4$ values encodes refined bounding-box positions for one of the $K$ classes.

\begin{figure}[t!]
\centering
\includegraphics[width=1\linewidth,trim=0 24em 25em 0, clip]{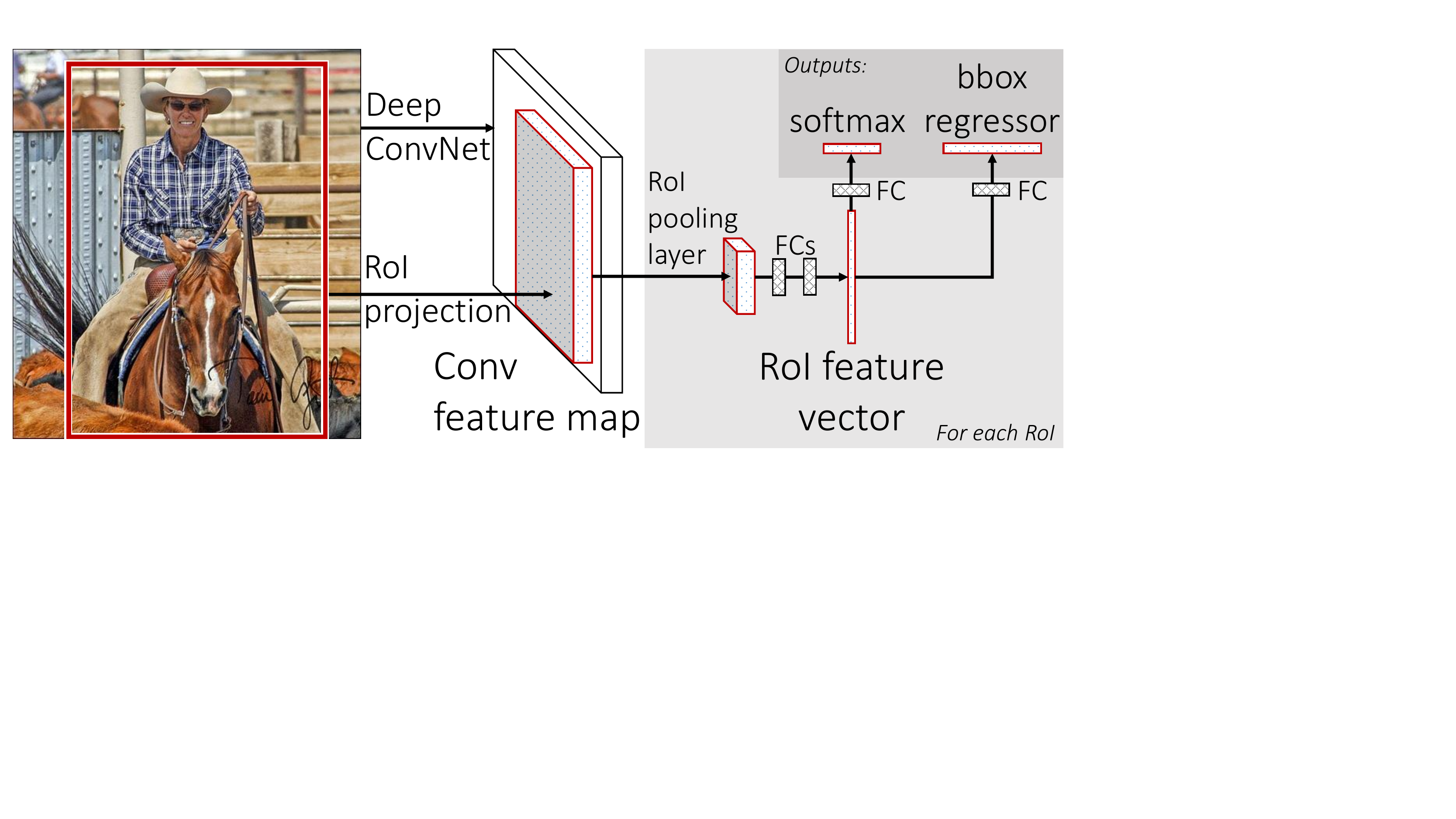}
%\vspace{-1em}
\caption{Fast R-CNN architecture. An input image and multiple regions of interest ({\roi}s) are input into a fully convolutional network. Each \roi is pooled into a fixed-size feature map and then mapped to a feature vector by fully connected layers (FCs). The network has two output vectors per \roi: softmax probabilities and per-class bounding-box regression offsets. The architecture is trained end-to-end with a multi-task loss.}
\figlabel{arch}
\end{figure}

\subsection{The \roi pooling layer}
The \roi pooling layer uses max pooling to convert the features inside any valid region of interest into a small feature map with a fixed spatial extent of $H \times W$ (\eg, $7 \times 7$), where $H$ and $W$ are layer hyper-parameters that are independent of any particular \roi.
In this paper, an \roi is a rectangular window into a conv feature map.
Each \roi is defined by a four-tuple $(r, c, h, w)$ that specifies its top-left corner $(r, c)$ and its height and width $(h, w)$.

\roi max pooling works by dividing the $h \times w$ RoI window into an $H \times W$ grid of sub-windows of approximate size $h / H \times w / W$ and then max-pooling the values in each sub-window into the corresponding output grid cell.
Pooling is applied independently to each feature map channel, as in standard max pooling.
The \roi layer is simply the special-case of the spatial pyramid pooling layer used in SPPnets \cite{he2014spp} in which there is only one pyramid level.
We use the pooling sub-window calculation given in \cite{he2014spp}.

%The region of interest (\roi) pooling layer is a simplified version of the spatial pyramid pooling used in SPPnet \cite{he2014spp}, in which our ``pyramid'' has only a single level.
%A \roi pooling layer takes as input $N$ feature maps and a list of $R$ regions of interest; typically $R \gg N$.
%The $N$ feature maps are supplied by the last conv layer of the network and each is a multi-dimensional array of size $H \times W \times C$, with $H$ rows, $W$ columns, and $C$ channels.
%Each \roi is a tuple $(n, r, c, h, w)$ that specifies a feature map index $n \in \{0, \ldots, N-1\}$, the \roi's top-left location $(r, c)$, and its height and width $(h, w)$.
%\roi pooling layers output max-pooled feature maps with spatial extent $H' \times W'$ and the original $C$ channels ($H' \le H$ and $W' \le W$).
%
%For each of the $R$ {\roi}s, the \roi pooling layer performs max pooling over $H'W'$ output bins.
%The bin sizes ($\approx h/H' \times w/W'$) are adaptively set such that they tile the $h \times w$ rectangle in the indexed feature map.
%We use the adaptive bin-size calculation given in \cite{he2014spp}.

\subsection{Initializing from pre-trained networks}
We experiment with three pre-trained ImageNet \cite{imagenet_cvpr09} networks, each with five max pooling layers and between five and thirteen conv layers (see \secref{setup} for network details).
When a pre-trained network initializes a Fast R-CNN network, it undergoes three transformations.

First, the last max pooling layer is replaced by a \roi pooling layer that is configured by setting $H$ and $W$ to be compatible with the net's first fully connected layer (\eg, $H = W = 7$ for \vggsixteen).

Second, the network's last fully connected layer and softmax (which were trained for 1000-way ImageNet classification) are replaced with the two sibling layers described earlier (a fully connected layer and softmax over $K + 1$ categories and category-specific bounding-box regressors).

Third, the network is modified to take two data inputs: a list of images and a list of {\roi}s in those images.
%Each \roi four-tuple is augmented with the index $n \in \{0,\ldots,N-1\}$ of its image in the input batch.
%Typical values used during training are $N = 2$ and $R = 128$.
%The number of {\roi}s can change dynamically from batch to batch, as can the resolution of the input images: all conv layers reshape on-the-fly in proportion to the input resolution \cite{lecun94}.
%and the \roi pooling layer produces fixed-size outputs.
%During training $N$ will be small (\eg, 2) and each batch comprises independently sampled images to be used in one SGD update.
%The $R$ regions of interest will be 128 {\roi}s sampled from the $N$ input images.
%During testing $N$ will either be 1, in the case of processing an image at a single scale, or a small number (\eg, 5) in the case that the test image is processed at multiple scales.

\subsection{Fine-tuning for detection}
Training all network weights with back-propagation is an important capability of Fast R-CNN.
First, let's elucidate why SPPnet is unable to update weights below the spatial pyramid pooling layer.

The root cause is that back-propagation through the SPP layer is highly inefficient when each training sample (\ie \roi) comes from a different image, which is exactly how R-CNN and SPPnet networks are trained.
The inefficiency stems from the fact that each \roi may have a very large receptive field, often spanning the entire input image.
Since the forward pass must process the entire receptive field, the training inputs are large (often the entire image).

We propose a more efficient training method that takes advantage of feature sharing during training.
In Fast R-CNN training, stochastic gradient descent (SGD) mini-batches are sampled hierarchically, first by sampling $N$ images and then by sampling $R/N$ {\roi}s from each image.
Critically, {\roi}s from the same image share computation and memory in the forward and backward passes.
Making $N$ small decreases mini-batch computation.
For example, when using $N = 2$ and $R = 128$, the proposed training scheme is roughly 64\X faster than sampling one {\roi} from $128$ different images (\ie, the R-CNN and SPPnet strategy).

One concern over this strategy is it may cause slow training convergence because {\roi}s from the same image are correlated.
This concern does not appear to be a practical issue and we achieve good results with $N = 2$ and $R = 128$ using fewer SGD iterations than R-CNN.

In addition to hierarchical sampling, Fast R-CNN uses a streamlined training process with one fine-tuning stage that jointly optimizes a softmax classifier and bounding-box regressors, rather than training a softmax classifier, SVMs, and regressors in three separate stages \cite{girshick2014rcnn,he2014spp}.
The components of this procedure (the loss, mini-batch sampling strategy, back-propagation through \roi pooling layers, and SGD hyper-parameters) are described below.

\paragraph{Multi-task loss.}
A Fast R-CNN network has two sibling output layers.
The first outputs a discrete probability distribution (per \roi), $p = (p_0,\ldots,p_K)$, over $K + 1$ categories.
As usual, $p$ is computed by a softmax over the $K + 1$ outputs of a fully connected layer.
The second sibling layer outputs bounding-box regression offsets, $t^{k} = \left(t^{k}_\textrm{x}, t^{k}_\textrm{y}, t^{k}_\textrm{w}, t^{k}_\textrm{h}\right)$, for each of the $K$ object classes, indexed by $k$.
We use the parameterization for $t^{k}$ given in \cite{girshick2014rcnn}, in which $t^k$ specifies a scale-invariant translation and log-space height/width shift relative to an object proposal.

Each training \roi is labeled with a ground-truth class $u$ and a ground-truth bounding-box regression target $v$.
We use a multi-task loss $L$ on each labeled {\roi} to jointly train for classification and bounding-box regression:
\begin{equation}
\eqlabel{loss}
L(p, u, t^u, v) = L_\textrm{cls}(p, u) + \lambda [u \ge 1] L_\textrm{loc}(t^u, v),
\end{equation}
in which $L_\textrm{cls}(p, u) = -\log p_u$ is log loss for true class $u$.

The second task loss, $L_{\textrm{loc}}$, is defined over a tuple of true bounding-box regression targets for class $u$, $v = (v_\textrm{x}, v_\textrm{y}, v_\textrm{w}, v_\textrm{h})$, and a predicted tuple $t^u = (t^u_\textrm{x}, t^u_\textrm{y}, t^u_\textrm{w}, t^u_\textrm{h})$, again for class $u$.
The Iverson bracket indicator function $[u \ge 1]$ evaluates to 1 when $u \ge 1$ and 0 otherwise.
By convention the catch-all background class is labeled $u = 0$.
For background {\roi}s there is no notion of a ground-truth bounding box and hence $L_\textrm{loc}$ is ignored.
For bounding-box regression, we use the loss
\begin{equation}
L_\textrm{loc}(t^u, v) = \sum_{i \in \{\textrm{x},\textrm{y},\textrm{w},\textrm{h}\}} \textrm{smooth}_{L_1}(t^u_i - v_i),
\end{equation}
in which
\begin{equation}
\eqlabel{smoothL1}
  \textrm{smooth}_{L_1}(x) =
  \begin{cases}
    0.5x^2& \text{if } |x| < 1\\
    |x| - 0.5& \text{otherwise},
  \end{cases}
\end{equation}
is a robust $L_1$ loss that is less sensitive to outliers than the $L_2$ loss used in R-CNN and SPPnet.
When the regression targets are unbounded, training with $L_2$ loss can require careful tuning of learning rates in order to prevent exploding gradients.
\eqref{smoothL1} eliminates this sensitivity.

The hyper-parameter $\lambda$ in \eqref{loss} controls the balance between the two task losses.
We normalize the ground-truth regression targets $v_i$ to have zero mean and unit variance.
All experiments use $\lambda = 1$.

We note that \cite{erhan2014scalable} uses a related loss to train a class-agnostic object proposal network.
Different from our approach, \cite{erhan2014scalable} advocates for a two-network system that separates localization and classification.
OverFeat \cite{overfeat}, R-CNN \cite{girshick2014rcnn}, and SPPnet \cite{he2014spp} also train classifiers and bounding-box localizers, however these methods use stage-wise training, which we show is suboptimal for Fast R-CNN (\secref{multitask}).

\paragraph{Mini-batch sampling.}
During fine-tuning, each SGD mini-batch is constructed from $N = 2$ images, chosen uniformly at random (as is common practice, we actually iterate over permutations of the dataset).
We use mini-batches of size $R = 128$, sampling $64$ {\roi}s from each image.
As in \cite{girshick2014rcnn}, we take 25\% of the {\roi}s from object proposals that have intersection over union (IoU) overlap with a ground-truth bounding box of at least $0.5$.
These {\roi}s comprise the examples labeled with a foreground object class, \ie $u \ge 1$.
%In cases of multiple overlap, the label comes from the ground-truth bounding box with which the sampled \roi has maximum IoU overlap.
The remaining {\roi}s are sampled from object proposals that have a maximum IoU with ground truth in the interval $[0.1, 0.5)$, following \cite{he2014spp}.
These are the background examples and are labeled with $u = 0$.
The lower threshold of $0.1$ appears to act as a heuristic for hard example mining \cite{lsvm-pami}.
During training, images are horizontally flipped with probability $0.5$.
No other data augmentation is used.%, though it would likely help.

\paragraph{Back-propagation through \roi pooling layers.}
%Fast R-CNN mini-batches start from whole images and hence contain all information needed to back-propagate derivatives from the loss function to the image.
%Back-propagation routes derivatives through the \roi pooling layer, as described below.
%Unlike SPPnet, we are able to fine-tuning all layers of the network because mini-batches start from whole images and hence posses the information needed to backpropagate derivatives from the loss to the image.
%Unlike SPPnet, we are able to fine-tune all layers of the network because mini-batches are constructed by first sampling images and then sampling {\roi}s within those images, rather than training from a set of pre-computed feature vectors.

%During the forward pass, an image list of $N = 2$ images is expanded by the \roi pooling layer into a mini-batch of size $R = 128$.
%The multi-task loss $L$ is averaged over the $R$ outputs.
%During back-propagation, derivatives flow through the \roi pooling layer.

Back-propagation routes derivatives through the \roi pooling layer.
For clarity, we assume only one image per mini-batch ($N=1$), though the extension to $N > 1$ is straightforward because the forward pass treats all images independently.

Let $x_i \in \mathbb{R}$ be the $i$-th activation input into the \roi pooling layer and let $y_{rj}$ be the layer's $j$-th output from the $r$-th \roi.
The \roi pooling layer computes $y_{rj} = x_{i^*(r,j)}$, in which $i^*(r,j) = \argmax_{i' \in \mathcal{R}(r,j)} x_{i'}$.
$\mathcal{R}(r,j)$ is the index set of inputs in the sub-window over which the output unit $y_{rj}$ max pools.
A single $x_i$ may be assigned to several different outputs $y_{rj}$.

The \roi pooling layer's \texttt{backwards} function computes partial derivative of the loss function with respect to each input variable $x_i$ by following the argmax switches:
\begin{equation}
  \frac{\partial L}{\partial x_i} = \sum_{r} \sum_{j}
  \left[i = i^*(r,j)\right] \frac{\partial L}{\partial y_{rj}}.
\end{equation}
In words, for each mini-batch \roi $r$ and for each pooling output unit $y_{rj}$, the partial derivative $\partial L/\partial y_{rj}$ is accumulated if $i$ is the argmax selected for $y_{rj}$ by max pooling.
In back-propagation, the partial derivatives $\partial L/\partial y_{rj}$ are already computed by the \texttt{backwards} function of the layer on top of the \roi pooling layer.

\paragraph{SGD hyper-parameters.}
The fully connected layers used for softmax classification and bounding-box regression are initialized from zero-mean Gaussian distributions with standard deviations $0.01$ and $0.001$, respectively.
Biases are initialized to $0$.
All layers use a per-layer learning rate of 1 for weights and 2 for biases and a global learning rate of $0.001$.
When training on VOC07 or VOC12 trainval we run SGD for 30k mini-batch iterations, and then lower the learning rate to $0.0001$ and train for another 10k iterations.
When we train on larger datasets, we run SGD for more iterations, as described later.
A momentum of $0.9$ and parameter decay of $0.0005$ (on weights and biases) are used.

%We note that it is not clear a priori that constructing mini-batches from a small number of images ($N = 2$) should work.
%However, we find that training is well behaved with this parameter choice, likely because the SGD momentum has a strong enough averaging effect.

\subsection{Scale invariance}
We explore two ways of achieving scale invariant object detection: (1) via ``brute force'' learning and (2) by using image pyramids.
These strategies follow the two approaches in \cite{he2014spp}.
In the brute-force approach, each image is processed at a pre-defined pixel size during both training and testing.
The network must directly learn scale-invariant object detection from the training data.

The multi-scale approach, in contrast, provides approximate scale-invariance to the network through an image pyramid.
At test-time, the image pyramid is used to approximately scale-normalize each object proposal.
During multi-scale training, we randomly sample a pyramid scale each time an image is sampled, following \cite{he2014spp}, as a form of data augmentation.
We experiment with multi-scale training for smaller networks only, due to GPU memory limits.

\section{Fast R-CNN detection}
Once a Fast R-CNN network is fine-tuned, detection amounts to little more than running a forward pass (assuming object proposals are pre-computed).
The network takes as input an image (or an image pyramid, encoded as a list of images) and a list of $R$ object proposals to score.
At test-time, $R$ is typically around $2000$, although we will consider cases in which it is larger ($\approx$ $45$k).
When using an image pyramid, each \roi is assigned to the scale such that the scaled \roi is closest to $224^2$ pixels in area \cite{he2014spp}.
%For single-scale detection, all {\roi}s index the single image in the input batch.
%In the multi-scale case, the input batch is an image pyramid.
%Here, each \roi specifies a pyramid level $l$ and a rectangle within that level.

For each test \roi $r$, the forward pass outputs a class posterior probability distribution $p$ and a set of predicted bounding-box offsets relative to $r$ (each of the $K$ classes gets its own refined bounding-box prediction).
We assign a detection confidence to $r$ for each object class $k$ using the estimated probability $\textrm{Pr}(\textrm{class} = k~|~r) \stackrel{\Delta}{=} p_k$.
We then perform non-maximum suppression independently for each class using the algorithm and settings from R-CNN \cite{girshick2014rcnn}.

\subsection{Truncated SVD for faster detection}
\seclabel{svd}
For whole-image classification, the time spent computing the fully connected layers is small compared to the conv layers.
On the contrary, for detection the number of {\roi}s to process is large and nearly half of the forward pass time is spent computing the fully connected layers (see \figref{timing}).
Large fully connected layers are easily accelerated by compressing them with truncated SVD \cite{Denton2014SVD,Xue2013svd}.

In this technique, a layer parameterized by the $u \times v$ weight matrix $W$ is approximately factorized as
\begin{equation}
  W \approx U \Sigma_t V^T
\end{equation}
using SVD.
In this factorization, $U$ is a $u \times t$ matrix comprising the first $t$ left-singular vectors of $W$, $\Sigma_t$ is a $t \times t$ diagonal matrix containing the top $t$ singular values of $W$, and $V$ is $v \times t$ matrix comprising the first $t$ right-singular vectors of $W$.
Truncated SVD reduces the parameter count from $uv$ to $t(u + v)$, which can be significant if $t$ is much smaller than $\min(u, v)$.
To compress a network, the single fully connected layer corresponding to $W$ is replaced by two fully connected layers, without a non-linearity between them.
The first of these layers uses the weight matrix $\Sigma_t V^T$ (and no biases) and the second uses $U$ (with the original biases associated with $W$).
This simple compression method gives good speedups when the number of {\roi}s is large.

%\subsection{Sum pooling and integral images}
%For networks with many feature channels (\eg $C = 512$), \roi pooling over a large number of regions also takes a non-trivial amount of time.
%If the max operation in \roi pooling is switched to either a sum or average, then integral images can be used to pool {\roi}s more quickly.
%\todo{run experiments}

\section{Main results}

\begin{table*}[t!]
\centering
\renewcommand{\arraystretch}{1.2}
\renewcommand{\tabcolsep}{1.2mm}
\resizebox{\linewidth}{!}{
  \begin{tabular}{@{}L{2.5cm}|L{1.2cm}|r*{19}{x}|x@{}}
method & train set & aero      & bike      & bird      & boat      & bottle     & bus        & car        & cat        & chair      & cow        & table      & dog        & horse      & mbike      & persn     & plant      & sheep      & sofa       & train      & tv         & mAP       \\
\hline
SPPnet BB \cite{he2014spp}$^\dagger$ &
07 $\setminus$ diff &
73.9 &
72.3 &
62.5 &
51.5 &
44.4 &
74.4 &
73.0 &
74.4 &
42.3 &
73.6 &
57.7 &
70.3 &
74.6 &
74.3 &
54.2 &
34.0 &
56.4 &
56.4 &
67.9 &
73.5 &
63.1 \\
R-CNN BB \cite{rcnn-pami} &
07 &
73.4 &
77.0 &
63.4 &
45.4 &
\bf{44.6} &
75.1 &
78.1 &
79.8 &
40.5 &
73.7 &
62.2 &
79.4 &
78.1 &
73.1 &
64.2 &
\bf{35.6} &
66.8 &
67.2 &
70.4 &
\bf{71.1} &
66.0 \\
\hline
FRCN [ours] &
07 &
74.5 &
78.3 &
69.2 &
53.2 &
36.6 &
77.3 &
78.2 &
82.0 &
40.7 &
72.7 &
67.9 &
79.6 &
79.2 &
73.0 &
69.0 &
30.1 &
65.4 &
70.2 &
75.8 &
65.8 &
66.9 \\
FRCN [ours] &
07 $\setminus$ diff &
74.6 &
\bf{79.0} &
68.6 &
57.0 &
39.3 &
79.5 &
\bf{78.6} &
81.9 &
\bf{48.0} &
74.0 &
67.4 &
80.5 &
80.7 &
74.1 &
69.6 &
31.8 &
67.1 &
68.4 &
75.3 &
65.5 &
68.1 \\
FRCN [ours] &
07+12 &
\bf{77.0} &
78.1 &
\bf{69.3} &
\bf{59.4} &
38.3 &
\bf{81.6} &
\bf{78.6} &
\bf{86.7} &
42.8 &
\bf{78.8} &
\bf{68.9} &
\bf{84.7} &
\bf{82.0} &
\bf{76.6} &
\bf{69.9} &
31.8 &
\bf{70.1} &
\bf{74.8} &
\bf{80.4} &
70.4 &
\bf{70.0} \\
\end{tabular}
}
\vspace{0.05em}
\caption{{\bf VOC 2007 test} detection average precision (\%). All methods use \vggsixteen. Training set key: {\bf 07}: VOC07 trainval, {\bf 07 $\setminus$ diff}: {\bf 07} without ``difficult'' examples, {\bf 07+12}: union of {\bf 07} and VOC12 trainval.
$^\dagger$SPPnet results were prepared by the authors of \cite{he2014spp}.}
\tablelabel{voc2007}
\end{table*}

\begin{table*}[t!]
\centering
\renewcommand{\arraystretch}{1.2}
\renewcommand{\tabcolsep}{1.2mm}
\resizebox{\linewidth}{!}{
\begin{tabular}{@{}L{2.5cm}|L{1.2cm}|r*{19}{x}|x@{}}
method & train set & aero      & bike      & bird      & boat      & bottle     & bus        & car        & cat        & chair      & cow        & table      & dog        & horse      & mbike      & persn     & plant      & sheep      & sofa       & train      & tv         & mAP       \\
\hline
BabyLearning &
Prop. &
77.7 &
73.8 &
62.3 &
48.8 &
45.4 &
67.3 &
67.0 &
80.3 &
41.3 &
70.8 &
49.7 &
79.5 &
74.7 &
78.6 &
64.5 &
36.0 &
69.9 &
55.7 &
70.4 &
61.7 &
63.8 \\
R-CNN BB \cite{rcnn-pami} &
12 &
79.3 &
72.4 &
63.1 &
44.0 &
44.4 &
64.6 &
66.3 &
84.9 &
38.8 &
67.3 &
48.4 &
82.3 &
75.0 &
76.7 &
65.7 &
35.8 &
66.2 &
54.8 &
69.1 &
58.8 &
62.9 \\
SegDeepM &
12+seg &
\bf{82.3} &
75.2 &
67.1 &
50.7 &
\bf{49.8} &
71.1 &
69.6 &
88.2 &
42.5 &
71.2 &
50.0 &
85.7 &
76.6 &
81.8 &
69.3 &
\bf{41.5} &
\bf{71.9} &
62.2 &
73.2 &
\bf{64.6} &
67.2 \\
\hline
FRCN [ours] &
12 &
80.1 &
74.4 &
67.7 &
49.4 &
41.4 &
74.2 &
68.8 &
87.8 &
41.9 &
70.1 &
50.2 &
86.1 &
77.3 &
81.1 &
70.4 &
33.3 &
67.0 &
63.3 &
77.2 &
60.0 &
66.1 \\
FRCN [ours] &
07++12 &
82.0 &
\bf{77.8} &
\bf{71.6} &
\bf{55.3} &
42.4 &
\bf{77.3} &
\bf{71.7} &
\bf{89.3} &
\bf{44.5} &
\bf{72.1} &
\bf{53.7} &
\bf{87.7} &
\bf{80.0} &
\bf{82.5} &
\bf{72.7} &
36.6 &
68.7 &
\bf{65.4} &
\bf{81.1} &
62.7 &
\bf{68.8} \\
\end{tabular}
}
\vspace{0.05em}
\caption{{\bf VOC 2010 test} detection average precision (\%).
BabyLearning uses a network based on \cite{Lin2014NiN}.
All other methods use \vggsixteen. Training set key: {\bf 12}: VOC12 trainval, {\bf Prop.}: proprietary dataset, {\bf 12+seg}: {\bf 12} with segmentation annotations, {\bf 07++12}: union of VOC07 trainval, VOC07 test, and VOC12 trainval.
%Results:
%$^\dagger$\url{http://host.robots.ox.ac.uk:8080/anonymous/UKA8XB.html},
%$^\dagger$\url{http://goo.gl/f1vtls},
%$^\ddagger$\url{http://host.robots.ox.ac.uk:8080/anonymous/DGTVGW.html}
%$^\ddagger$\url{http://goo.gl/kyZcnW}
}
\tablelabel{voc2010}
\end{table*}

\begin{table*}[t!]
\centering
\renewcommand{\arraystretch}{1.2}
\renewcommand{\tabcolsep}{1.2mm}
\resizebox{\linewidth}{!}{
  \begin{tabular}{@{}L{2.5cm}|L{1.2cm}|r*{19}{x}|x@{}}
method & train set & aero      & bike      & bird      & boat      & bottle     & bus        & car        & cat        & chair      & cow        & table      & dog        & horse      & mbike      & persn     & plant      & sheep      & sofa       & train      & tv         & mAP       \\
\hline
BabyLearning &
Prop. &
78.0 &
74.2 &
61.3 &
45.7 &
42.7 &
68.2 &
66.8 &
80.2 &
40.6 &
70.0 &
49.8 &
79.0 &
74.5 &
77.9 &
64.0 &
35.3 &
67.9 &
55.7 &
68.7 &
62.6 &
63.2 \\
NUS\_NIN\_c2000 &
Unk. &
80.2 &
73.8 &
61.9 &
43.7 &
\bf{43.0} &
70.3 &
67.6 &
80.7 &
41.9 &
69.7 &
51.7 &
78.2 &
75.2 &
76.9 &
65.1 &
\bf{38.6} &
\bf{68.3} &
58.0 &
68.7 &
63.3 &
63.8 \\
R-CNN BB \cite{rcnn-pami} &
12 &
79.6 &
72.7 &
61.9 &
41.2 &
41.9 &
65.9 &
66.4 &
84.6 &
38.5 &
67.2 &
46.7 &
82.0 &
74.8 &
76.0 &
65.2 &
35.6 &
65.4 &
54.2 &
67.4 &
60.3 &
62.4 \\
\hline
FRCN [ours] &
12 &
80.3 &
74.7 &
66.9 &
46.9 &
37.7 &
73.9 &
68.6 &
87.7 &
41.7 &
71.1 &
51.1 &
86.0 &
77.8 &
79.8 &
69.8 &
32.1 &
65.5 &
63.8 &
76.4 &
61.7 &
65.7 \\
FRCN [ours] &
07++12 &
\bf{82.3} &
\bf{78.4} &
\bf{70.8} &
\bf{52.3} &
38.7 &
\bf{77.8} &
\bf{71.6} &
\bf{89.3} &
\bf{44.2} &
\bf{73.0} &
\bf{55.0} &
\bf{87.5} &
\bf{80.5} &
\bf{80.8} &
\bf{72.0} &
35.1 &
\bf{68.3} &
\bf{65.7} &
\bf{80.4} &
\bf{64.2} &
\bf{68.4} \\
\end{tabular}
}
\vspace{0.05em}
\caption{{\bf VOC 2012 test} detection average precision (\%).
BabyLearning and NUS\_NIN\_c2000 use networks based on \cite{Lin2014NiN}.
All other methods use \vggsixteen. Training set key: see \tableref{voc2010}, {\bf Unk.}: unknown.
%Results:
%$^\dagger$\url{http://host.robots.ox.ac.uk:8080/anonymous/QMSIIY.html},
%$^\ddagger$\url{http://host.robots.ox.ac.uk:8080/anonymous/DCSBZY.html}
%$^\dagger$\url{http://goo.gl/weNq2Z},
%$^\ddagger$\url{http://goo.gl/o1tZ10}
}
\tablelabel{voc2012}
\end{table*}

Three main results support this paper's contributions:
\begin{enumerate}
  \itemsep0em
  \item State-of-the-art mAP on VOC07, 2010, and 2012
  \item Fast training and testing compared to R-CNN, SPPnet
  \item Fine-tuning conv layers in \vggsixteen improves mAP
\end{enumerate}

\subsection{Experimental setup}
\seclabel{setup}
Our experiments use three pre-trained ImageNet models that are available online.\footnote{\url{https://github.com/BVLC/caffe/wiki/Model-Zoo}}
The first is the CaffeNet (essentially AlexNet \cite{krizhevsky2012imagenet}) from R-CNN \cite{girshick2014rcnn}.
We alternatively refer to this CaffeNet as model \Sm, for ``small.''
The second network is VGG\_CNN\_M\_1024 from \cite{Chatfield14}, which has the same depth as \Sm, but is wider.
We call this network model \Med, for ``medium.''
The final network is the very deep \vggsixteen model from \cite{simonyan2015verydeep}.
Since this model is the largest, we call it model \Lg.
In this section, all experiments use \emph{single-scale} training and testing ($s = 600$; see \secref{scale} for details).

\subsection{VOC 2010 and 2012 results}
On these datasets, we compare Fast R-CNN (\emph{FRCN}, for short) against the top methods on the \texttt{comp4} (outside data) track from the public leaderboard (\tableref{voc2010}, \tableref{voc2012}).\footnote{\url{http://host.robots.ox.ac.uk:8080/leaderboard} (accessed April 18, 2015)}
For the NUS\_NIN\_c2000 and BabyLearning methods, there are no associated publications at this time and we could not find exact information on the ConvNet architectures used; they are variants of the Network-in-Network design \cite{Lin2014NiN}.
All other methods are initialized from the same pre-trained \vggsixteen network.
%We show results when training on the VOC12 trainval image set as well as an augmented dataset that includes all annotated images in VOC07.

Fast R-CNN achieves the top result on VOC12 with a mAP of 65.7\% (and 68.4\% with extra data).
It is also two orders of magnitude faster than the other methods, which are all based on the ``slow'' R-CNN pipeline.
On VOC10, SegDeepM \cite{Zhu2015segDeepM} achieves a higher mAP than Fast R-CNN (67.2\% vs. 66.1\%).
SegDeepM is trained on VOC12 trainval plus segmentation annotations; it is designed to boost R-CNN accuracy by using a Markov random field to reason over R-CNN detections and segmentations from the O$_2$P \cite{o2p} semantic-segmentation method.
Fast R-CNN can be swapped into SegDeepM in place of R-CNN, which may lead to better results.
When using the enlarged 07++12 training set (see \tableref{voc2010} caption), Fast R-CNN's mAP increases to 68.8\%, surpassing SegDeepM.

\subsection{VOC 2007 results}
On VOC07, we compare Fast R-CNN to R-CNN and SPPnet.
All methods start from the same pre-trained \vggsixteen network and use bounding-box regression.
The \vggsixteen SPPnet results were computed by the authors of \cite{he2014spp}.
SPPnet uses five scales during both training and testing.
The improvement of Fast R-CNN over SPPnet illustrates that even though Fast R-CNN uses single-scale training and testing, fine-tuning the conv layers provides a large improvement in mAP (from 63.1\% to 66.9\%).
%Moreover, single-scale testing significantly speeds up detection.
R-CNN achieves a mAP of 66.0\%.
%Fast R-CNN surpasses R-CNN, at 66.9\%.
%These results are pragmatically valuable given how much faster and easier Fast R-CNN is to train and test, which we discuss next.
As a minor point, SPPnet was trained \emph{without} examples marked as ``difficult'' in PASCAL.
Removing these examples improves Fast R-CNN mAP to 68.1\%.
All other experiments use ``difficult'' examples.

\subsection{Training and testing time}
% Show how long R-CNN and SPP-net take to train and test on VOC07.
Fast training and testing times are our second main result.
\tableref{timing} compares training time (hours), testing rate (seconds per image), and mAP on VOC07 between Fast R-CNN, R-CNN, and SPPnet.
For \vggsixteen, Fast R-CNN processes images 146\X faster than R-CNN without truncated SVD and 213\X faster with it.
Training time is reduced by 9\X, from 84 hours to 9.5.
Compared to SPPnet, Fast R-CNN trains \vggsixteen 2.7\X faster (in 9.5 vs. 25.5 hours) and tests 7\X faster without truncated SVD or 10\X faster with it.
Fast R-CNN also eliminates hundreds of gigabytes of disk storage, because it does not cache features.

\begin{table}[h!]
\begin{center}
\setlength{\tabcolsep}{3pt}
\renewcommand{\arraystretch}{1.2}
\resizebox{\linewidth}{!}{
\small
\begin{tabular}{l|rrr|rrr|r}
  & \multicolumn{3}{c|}{Fast R-CNN} & \multicolumn{3}{c|}{R-CNN} & \multicolumn{1}{c}{SPPnet} \\
  & \Sm & \Med & \Lg & \Sm & \Med & \Lg & $^\dagger$\Lg \\
%  & \multicolumn{3}{c|}{\Sm} & \multicolumn{3}{c|}{\Med} & \multicolumn{3}{c}{\Lg}  \\
\hline
train time (h) & \bf{1.2} & 2.0 & 9.5 &
22 & 28 & 84 & 25 \\
train speedup & \bf{18.3\X} & 14.0\X & 8.8\X &
1\X & 1\X & 1\X & 3.4\X \\
\hline
test rate (s/im) & 0.10 & 0.15 & 0.32 &
9.8 & 12.1 & 47.0 & 2.3 \\
~$\rhd$ with SVD & \bf{0.06} & 0.08 & 0.22 &
- & - & - & - \\
\hline
test speedup & 98\X & 80\X & 146\X &
1\X & 1\X & 1\X & 20\X \\
~$\rhd$ with SVD & 169\X & 150\X & \bf{213\X} &
- & - & - & - \\
\hline
VOC07 mAP & 57.1 & 59.2 & \bf{66.9} &
58.5 & 60.2 & 66.0 & 63.1 \\
~$\rhd$ with SVD & 56.5 & 58.7 & 66.6 &
- & - & - & - \\
\end{tabular}
}
\end{center}
\caption{Runtime comparison between the same models in Fast R-CNN, R-CNN, and SPPnet.
Fast R-CNN uses single-scale mode.
SPPnet uses the five scales specified in \cite{he2014spp}.
$^\dagger$Timing provided by the authors of \cite{he2014spp}.
Times were measured on an Nvidia K40 GPU.
}
\tablelabel{timing}
\vspace{-1em}
\end{table}

\paragraph{Truncated SVD.}
%\tableref{timing} includes results with and without truncated SVD acceleration.
%For image classification, truncated SVD is primarily useful as a model compression technique to reduce storage space.
%But for detection, it also provides a significant speed-up.
Truncated SVD can reduce detection time by more than 30\% with only a small (0.3 percentage point) drop in mAP and without needing to perform additional fine-tuning after model compression.
\figref{timing} illustrates how using the top $1024$ singular values from the $25088 \times 4096$ matrix in {\vggsixteen}'s fc6 layer and the top $256$ singular values from the $4096 \times 4096$ fc7 layer reduces runtime with little loss in mAP.
Further speed-ups are possible with smaller drops in mAP if one fine-tunes again after compression.
%Here we focus on keeping training as simple as possible.

\begin{figure}[h!]
\centering
\includegraphics[width=0.49\linewidth,trim=3em 2em 0 0, clip]{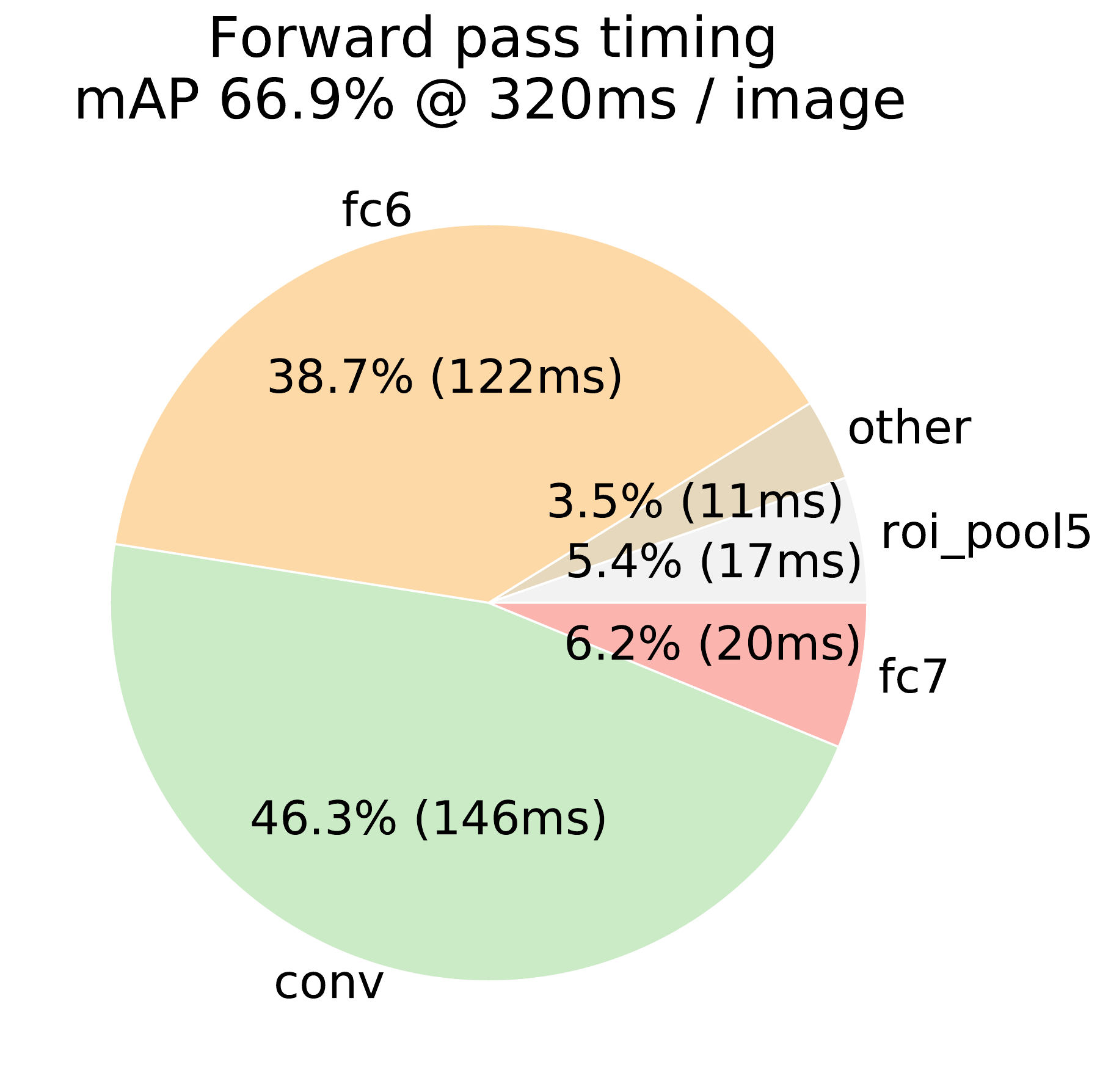}
\includegraphics[width=0.49\linewidth,trim=3em 2em 0 0, clip]{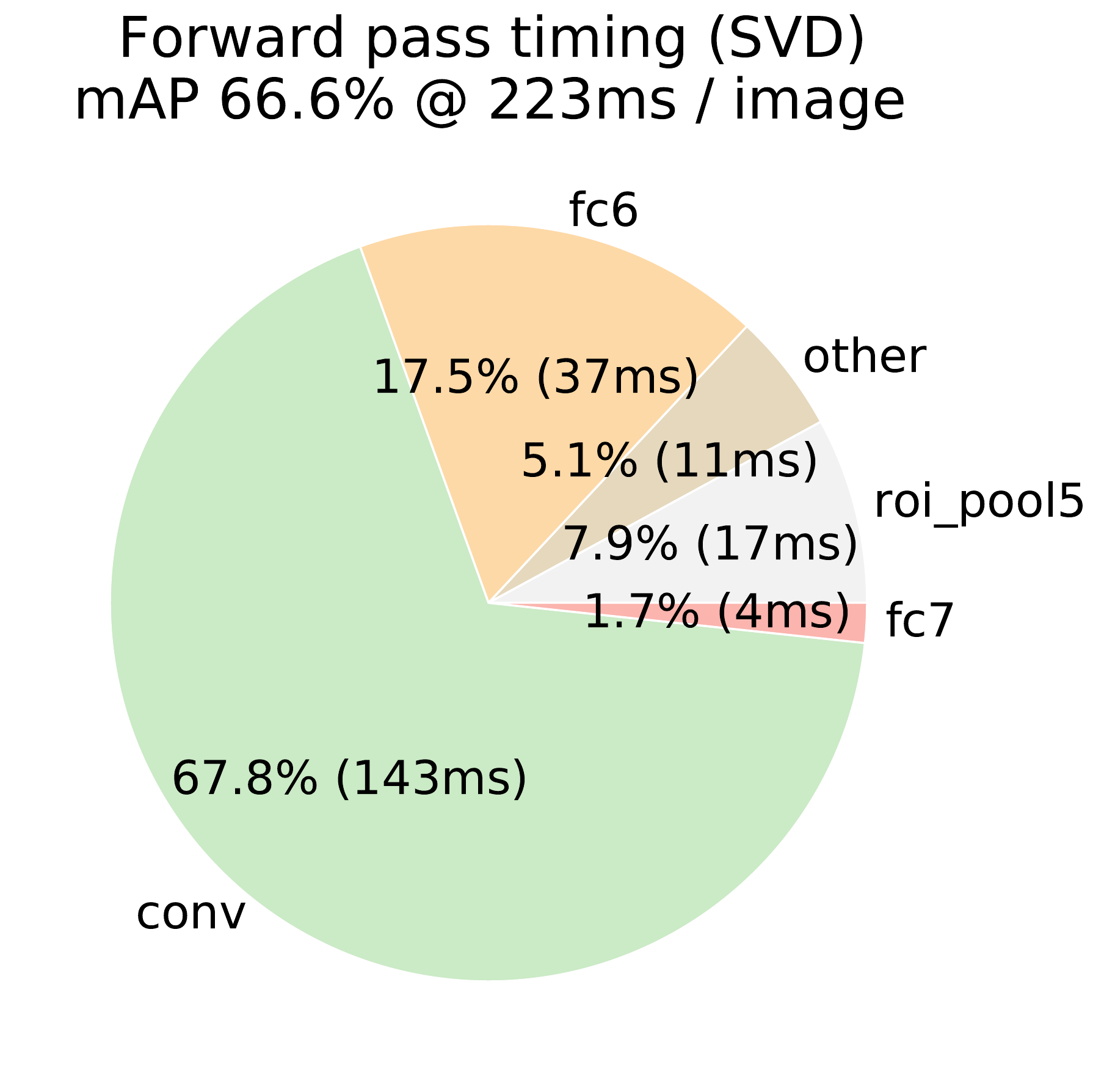}
%\vspace{-1em}
\caption{Timing for \vggsixteen before and after truncated SVD.
Before SVD, fully connected layers fc6 and fc7 take 45\% of the time.}
\figlabel{timing}
\end{figure}

%\begin{table}[h!]
%\begin{center}
%\setlength{\tabcolsep}{6pt}
%\renewcommand{\arraystretch}{1.1}
%\small
%\begin{tabular}{l|rr|rr|rr}
%  & \multicolumn{2}{c|}{\Sm} & \multicolumn{2}{c|}{\Med} & \multicolumn{2}{c}{\Lg} \\
%\hline
%Trunc. SVD? &  & \checkmark &  & \checkmark & & \checkmark \\
%s / image & 0.10 & 0.06 & 0.15 & 0.08 & 0.32 & 0.22 \\
%VOC07 mAP & 57.1 & 56.5 & 59.2 & 58.7 & 66.9 & 66.6
%\end{tabular}
%\end{center}
%\caption{Effects of truncated SVD on test speed and mAP.}
%\tablelabel{svd}
%\end{table}
%
%\tableref{svd} shows results for all three models using the top 1024 and 256 singular values for layers fc6 and fc7, respectively.
%Even though the compression factor increases from \Sm to \Lg, the loss in mAP decreases.

\subsection{Which layers to fine-tune?}
For the less deep networks considered in the SPPnet paper \cite{he2014spp}, fine-tuning only the fully connected layers appeared to be sufficient for good accuracy.
We hypothesized that this result would not hold for very deep networks.
To validate that fine-tuning the conv layers is important for \vggsixteen, we use Fast R-CNN to fine-tune, but \emph{freeze} the thirteen conv layers so that only the fully connected layers learn.
This ablation emulates single-scale SPPnet training and \emph{decreases mAP from 66.9\% to 61.4\%} (\tableref{whichlayers}).
This experiment verifies our hypothesis: training through the \roi pooling layer is important for very deep nets.

\begin{table}[h!]
\begin{center}
\setlength{\tabcolsep}{3pt}
\renewcommand{\arraystretch}{1.1}
\small
\begin{tabular}{l|rrr|r}
  & \multicolumn{3}{c|}{layers that are fine-tuned in model \Lg} & SPPnet \Lg \\
  & $\ge$ fc6 & $\ge$ conv3\_1 & $\ge$ conv2\_1 & $\ge$ fc6 \\
\hline
VOC07 mAP & 61.4 & 66.9 & \bf{67.2} & 63.1 \\
test rate (s/im) & \bf{0.32} & \bf{0.32} & \bf{0.32} & 2.3 \\
\end{tabular}
\end{center}
\caption{Effect of restricting which layers are fine-tuned for \vggsixteen.
Fine-tuning $\ge$ fc6 emulates the SPPnet training algorithm \cite{he2014spp}, but using a single scale.
SPPnet \Lg results were obtained using five scales, at a significant (7\X) speed cost.}
\tablelabel{whichlayers}
\vspace{-0.5em}
\end{table}

Does this mean that \emph{all} conv layers should be fine-tuned? In short, \emph{no}.
In the smaller networks (\Sm and \Med) we find that conv1 is generic and task independent (a well-known fact \cite{krizhevsky2012imagenet}).
Allowing conv1 to learn, or not, has no meaningful effect on mAP.
For \vggsixteen, we found it only necessary to update layers from conv3\_1 and up (9 of the 13 conv layers).
This observation is pragmatic: (1) updating from conv2\_1 slows training by 1.3\X (12.5 vs. 9.5 hours) compared to learning from conv3\_1;
and (2) updating from conv1\_1 over-runs GPU memory.
The difference in mAP when learning from conv2\_1 up was only $+0.3$ points (\tableref{whichlayers}, last column).
All Fast R-CNN results in this paper using \vggsixteen fine-tune layers conv3\_1 and up; all experiments with models \Sm and \Med fine-tune layers conv2 and up.

\begin{table*}[t!]
\begin{center}
\setlength{\tabcolsep}{5pt}
\renewcommand{\arraystretch}{1.1}
\small
\begin{tabular}{l|rrrr|rrrr|rrrr}
  & \multicolumn{4}{c|}{\Sm} & \multicolumn{4}{c|}{\Med} & \multicolumn{4}{c}{\Lg}  \\
\hline
multi-task training? &
&
\checkmark &
&
\checkmark &
&
\checkmark &
&
\checkmark &
&
\checkmark &
&
\checkmark
\\
stage-wise training? &
 &
 &
\checkmark &
&
 &
 &
\checkmark &
&
 &
 &
\checkmark &
\\
test-time bbox reg? & & & \checkmark & \checkmark & & & \checkmark & \checkmark & & & \checkmark & \checkmark \\
VOC07 mAP & 52.2 & 53.3 & 54.6 & \bf{57.1} & 54.7 & 55.5 & 56.6 & \bf{59.2} & 62.6 & 63.4 & 64.0 & \bf{66.9} \\
\end{tabular}
\end{center}
\caption{Multi-task training (forth column per group) improves mAP over piecewise training (third column per group).}
\tablelabel{multitask}
\vspace{-0.5em}
\end{table*}

\section{Design evaluation}

We conducted experiments to understand how Fast R-CNN compares to R-CNN and SPPnet, as well as to evaluate design decisions.
Following best practices, we performed these experiments on the PASCAL VOC07 dataset.
%Based on those results, we train and test a couple of models on VOC12.

\subsection{Does multi-task training help?}
\seclabel{multitask}
Multi-task training is convenient because it avoids managing a pipeline of sequentially-trained tasks.
But it also has the potential to improve results because the tasks influence each other through a shared representation (the ConvNet) \cite{caruana1997multitask}.
Does multi-task training improve object detection accuracy in Fast R-CNN?

To test this question, we train baseline networks that use only the classification loss, $L_\textrm{cls}$, in \eqref{loss} (\ie, setting $\lambda = 0$).
These baselines are printed for models \Sm, \Med, and \Lg in the first column of each group in \tableref{multitask}.
Note that these models \emph{do not} have bounding-box regressors.
Next (second column per group), we take networks that were trained with the multi-task loss (\eqref{loss}, $\lambda = 1$), but we \emph{disable} bounding-box regression at test time.
This isolates the networks' classification accuracy and allows an apples-to-apples comparison with the baseline networks.
%(Results with bounding-box regression are shown in the third column, per group, to illustrate its effect.)

Across all three networks we observe that multi-task training improves pure classification accuracy relative to training for classification alone.
The improvement ranges from $+0.8$ to $+1.1$ mAP points, showing a consistent positive effect from multi-task learning.

Finally, we take the baseline models (trained with only the classification loss), tack on the bounding-box regression layer, and train them with $L_{loc}$ while keeping all other network parameters frozen.
The third column in each group shows the results of this \emph{stage-wise} training scheme: mAP improves over column one, but stage-wise training underperforms multi-task training (forth column per group).

\subsection{Scale invariance: to brute force or finesse?}
\seclabel{scale}
We compare two strategies for achieving scale-invariant object detection: brute-force learning (single scale) and image pyramids (multi-scale).
In either case, we define the scale $s$ of an image to be the length of its \emph{shortest} side.

All single-scale experiments use $s = 600$ pixels;
$s$ may be less than $600$ for some images as we cap the longest image side at $1000$ pixels and maintain the image's aspect ratio.
These values were selected so that \vggsixteen fits in GPU memory during fine-tuning.
The smaller models are not memory bound and can benefit from larger values of $s$; however, optimizing $s$ for each model is not our main concern.
We note that PASCAL images are $384 \times 473$ pixels on average and thus the single-scale setting typically upsamples images by a factor of 1.6.
The average effective stride at the \roi pooling layer is thus $\approx 10$ pixels.

In the multi-scale setting, we use the same five scales specified in \cite{he2014spp} ($s \in \{480, 576, 688, 864, 1200\}$) to facilitate comparison with SPPnet.
However, we cap the longest side at $2000$ pixels to avoid exceeding GPU memory.

\begin{table}[h!]
\begin{center}
\setlength{\tabcolsep}{4.7pt}
\renewcommand{\arraystretch}{1.1}
\small
\begin{tabular}{l|rr|rr|rr|r}
 & \multicolumn{2}{c|}{SPPnet \ZF}  & \multicolumn{2}{c|}{\Sm} & \multicolumn{2}{c|}{\Med} & \Lg \\
\hline
scales & 1 & 5 & 1 & 5 & 1 & 5 & 1 \\
test rate (s/im) & 0.14 & 0.38 & \bf{0.10} & 0.39 & 0.15 & 0.64 & 0.32 \\
VOC07 mAP & 58.0 & 59.2 & 57.1 & 58.4 & 59.2 & 60.7 & \bf{66.9}
\end{tabular}
\end{center}
\caption{Multi-scale vs. single scale.
SPPnet \ZF (similar to model \Sm) results are from \cite{he2014spp}.
Larger networks with a single-scale offer the best speed / accuracy tradeoff.
(\Lg cannot use multi-scale in our implementation due to GPU memory constraints.)
}
\tablelabel{scales}
\vspace{-0.5em}
\end{table}

\tableref{scales} shows models \Sm and \Med when trained and tested with either one or five scales.
Perhaps the most surprising result in \cite{he2014spp} was that single-scale detection performs almost as well as multi-scale detection.
Our findings confirm their result: deep ConvNets are adept at directly learning scale invariance.
The multi-scale approach offers only a small increase in mAP at a large cost in compute time (\tableref{scales}).
In the case of \vggsixteen (model \Lg), we are limited to using a single scale by implementation details. Yet it achieves a mAP of 66.9\%, which is slightly higher than the 66.0\% reported for R-CNN \cite{rcnn-pami}, even though R-CNN uses ``infinite'' scales in the sense that each proposal is warped to a canonical size.

Since single-scale processing offers the best tradeoff between speed and accuracy, especially for very deep models, all experiments outside of this sub-section use single-scale training and testing with $s = 600$ pixels.

\subsection{Do we need more training data?}
\seclabel{moredata}
A good object detector should improve when supplied with more training data.
Zhu \etal \cite{devaMoreData} found that DPM \cite{lsvm-pami} mAP saturates after only a few hundred to thousand training examples.
Here we augment the VOC07 trainval set with the VOC12 trainval set, roughly tripling the number of images to 16.5k, to evaluate Fast R-CNN.
%\cite{agrawal2014analyzing}.
Enlarging the training set improves mAP on VOC07 test from 66.9\% to 70.0\% (\tableref{voc2007}).
When training on this dataset we use 60k mini-batch iterations instead of 40k.

We perform similar experiments for VOC10 and 2012, for which we construct a dataset of 21.5k images from the union of VOC07 trainval, test, and VOC12 trainval.
When training on this dataset, we use 100k SGD iterations and lower the learning rate by $0.1\times$ each 40k iterations (instead of each 30k).
For VOC10 and 2012, mAP improves from 66.1\% to 68.8\% and from 65.7\% to 68.4\%, respectively.
%Fast R-CNN accuracy should improve if more training data become available.

\subsection{Do SVMs outperform softmax?}
Fast R-CNN uses the softmax classifier learnt during fine-tuning instead of training one-vs-rest linear SVMs post-hoc, as was done in R-CNN and SPPnet.
To understand the impact of this choice, we implemented post-hoc SVM training with hard negative mining in Fast R-CNN.
We use the same training algorithm and hyper-parameters as in R-CNN.
\begin{table}[h!]
\begin{center}
\setlength{\tabcolsep}{6pt}
\renewcommand{\arraystretch}{1.1}
\small
\begin{tabular}{l|l|r|r|r}
  method & classifier & \Sm & \Med & \Lg \\
\hline
R-CNN \cite{girshick2014rcnn,rcnn-pami} & SVM & \bf{58.5} & \bf{60.2} & 66.0 \\
\hline
FRCN [ours] & SVM & 56.3 & 58.7 & 66.8 \\
FRCN [ours] & softmax & 57.1 & 59.2 & \bf{66.9} \\
\end{tabular}
\end{center}
\caption{Fast R-CNN with softmax vs. SVM (VOC07 mAP).}
\tablelabel{svm}
\vspace{-0.5em}
\end{table}

\tableref{svm} shows softmax slightly outperforming SVM for all three networks, by $+0.1$ to $+0.8$ mAP points.
This effect is small, but it demonstrates that ``one-shot'' fine-tuning is sufficient compared to previous multi-stage training approaches.
We note that softmax, unlike one-vs-rest SVMs, introduces competition between classes when scoring a \roi.

\subsection{Are more proposals always better?}

%Fast R-CNN enables researchers to test ideas that would have been prohibitively slow in the past.
%For example, we can study the behavior of Fast R-CNN when using a large number of object proposals.

There are (broadly) two types of object detectors: those that use a \emph{sparse} set of object proposals (\eg, selective search \cite{UijlingsIJCV2013}) and those that use a \emph{dense} set (\eg, DPM \cite{lsvm-pami}).
Classifying sparse proposals is a type of \emph{cascade} \cite{Viola01} in which the proposal mechanism first rejects a vast number of candidates leaving the classifier with a small set to evaluate.
This cascade improves detection accuracy when applied to DPM detections \cite{UijlingsIJCV2013}.
We find evidence that the proposal-classifier cascade also improves Fast R-CNN accuracy.

%\paragraph{More sparse proposals.}
Using selective search's \emph{quality mode}, we sweep from 1k to 10k proposals per image, each time \emph{re-training} and \emph{re-testing} model \Med.
If proposals serve a purely computational role, increasing the number of proposals per image should not harm mAP.
\begin{figure}[h!]
\centering
\includegraphics[width=1\linewidth,trim=0em 0em 0 0, clip]{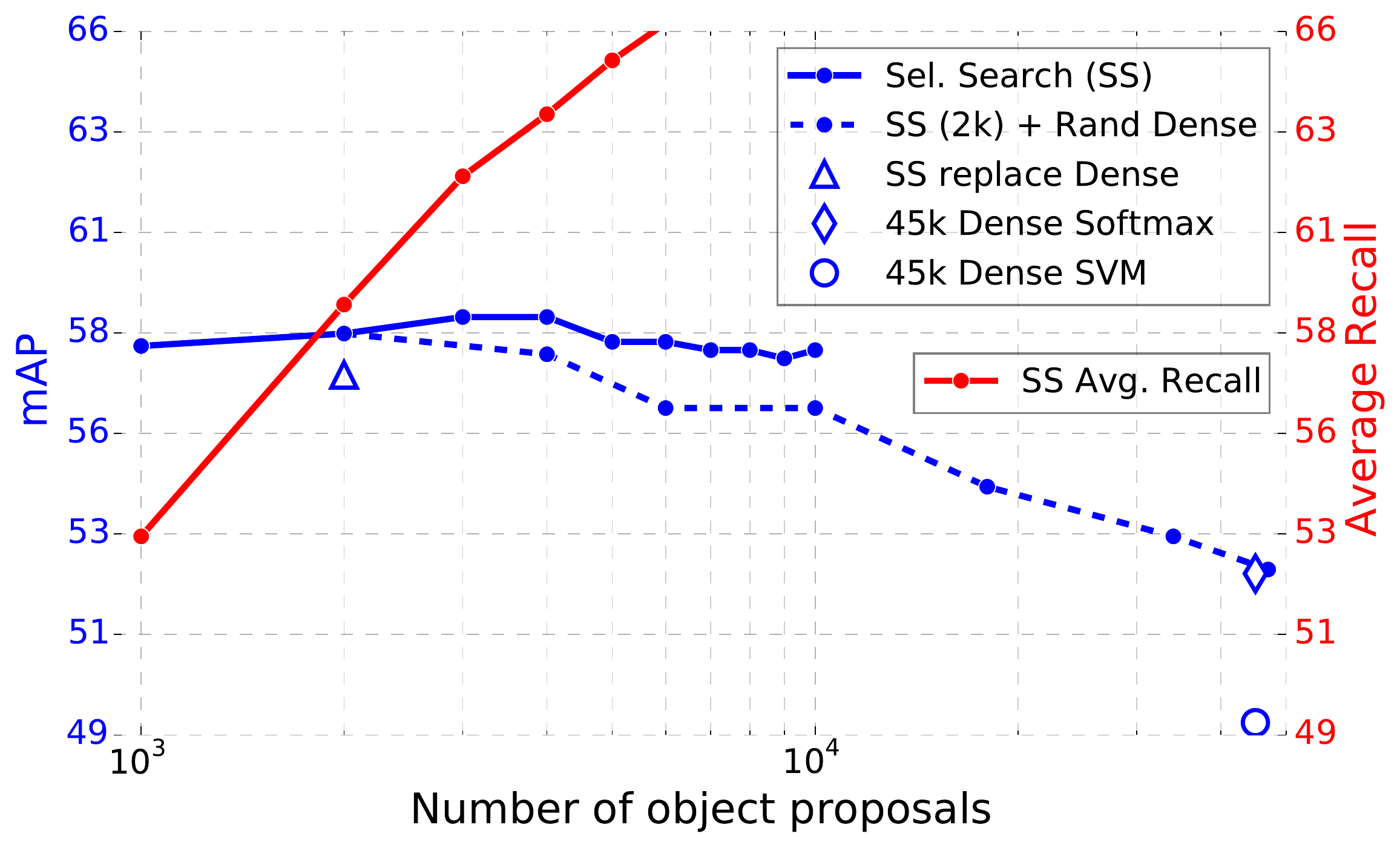}
%\vspace{-1em}
\caption{VOC07 test mAP and AR for various proposal schemes.}
\figlabel{proposals}
\end{figure}

We find that  mAP rises and then falls slightly as the proposal count increases (\figref{proposals}, solid blue line).
This experiment shows that swamping the deep classifier with more proposals does not help, and even slightly hurts, accuracy.

This result is difficult to predict without actually running the experiment.
The state-of-the-art for measuring object proposal quality is Average Recall (AR) \cite{Hosang15proposals}.
AR correlates well with mAP for several proposal methods using R-CNN, \emph{when using a fixed number of proposals per image}.
\figref{proposals} shows that AR (solid red line) does not correlate well with mAP as the number of proposals per image is varied.
AR must be used with care; higher AR due to more proposals does not imply that mAP will increase.
Fortunately, training and testing with model \Med takes less than 2.5 hours.
Fast R-CNN thus enables efficient, direct evaluation of object proposal mAP, which is preferable to proxy metrics.

%\paragraph{Dense proposals.}
We also investigate Fast R-CNN when using \emph{densely} generated boxes (over scale, position, and aspect ratio), at a rate of about 45k boxes / image.
This dense set is rich enough that when each selective search box is replaced by its closest (in IoU) dense box, mAP drops only 1 point (to 57.7\%, \figref{proposals}, blue triangle).

The statistics of the dense boxes differ from those of selective search boxes.
Starting with 2k selective search boxes, we test mAP when \emph{adding} a random sample of $1000 \times \{2,4,6,8,10,32,45\}$ dense boxes.
For each experiment we re-train and re-test model \Med.
When these dense boxes are added, mAP falls more strongly than when adding more selective search boxes, eventually reaching 53.0\%.

We also train and test Fast R-CNN using \emph{only} dense boxes (45k / image).
This setting yields a mAP of 52.9\% (blue diamond).
Finally, we check if SVMs with hard negative mining are needed to cope with the dense box distribution.
SVMs do even worse: 49.3\% (blue circle).

%\paragraph{Dense vs. sparse.}
%Sparse object proposal methods are currently the speed bottleneck in Fast R-CNN.
%%Selective search takes 2s / image and EdgeBoxes takes 0.2s / image.
%Replacing them with a dense set of ``sliding windows'' is attractive, since it is essentially free.
%Yet, these experiments provide the first evidence that sparse proposals do indeed ``improve detection quality by reducing spurious false positives'' \cite{Hosang15proposals}.

\subsection{Preliminary MS COCO results}
We applied Fast R-CNN (with \vggsixteen) to the MS COCO dataset \cite{coco} to establish a preliminary baseline.
We trained on the 80k image training set for 240k iterations and evaluated on the ``test-dev'' set using the evaluation server.
The PASCAL-style mAP is 35.9\%; the new COCO-style AP, which also averages over IoU thresholds, is 19.7\%.

\section{Conclusion}

This paper proposes Fast R-CNN, a clean and fast update to R-CNN and SPPnet.
In addition to reporting state-of-the-art detection results, we present detailed experiments that we hope provide new insights.
Of particular note, sparse object proposals appear to improve detector quality.
This issue was too costly (in time) to probe in the past, but becomes practical with Fast R-CNN.
Of course, there may exist yet undiscovered techniques that allow dense boxes to perform as well as sparse proposals.
Such methods, if developed, may help further accelerate object detection.

\paragraph{Acknowledgements.}
I thank Kaiming He, Larry Zitnick, and Piotr Doll{\'a}r for helpful discussions and encouragement.

{\small
\bibliographystyle{ieee}
\bibliography{main}
}

\end{document}